\documentclass[twoside,leqno,twocolumn]{article}

\usepackage[letterpaper]{geometry}

\usepackage{ltexpprt}
\usepackage{hyperref}
\usepackage{amsmath,amssymb,amsfonts}
\usepackage{mathtools}

\usepackage{yhmath}
\usepackage{hyperref}
\usepackage{url}
\usepackage{graphicx}
\usepackage{algorithm}
\usepackage{algpseudocode}
\usepackage{multirow}
\usepackage[switch]{lineno}
\usepackage{xurl}
\usepackage{makecell}

\usepackage{xcolor}

\begin{document}

\title{\Large Combining Satellite and Weather Data for Crop Type Mapping: An Inverse Modelling Approach}

\author{
    Praveen Ravirathinam$^\dag$\and
    Rahul Ghosh$^{\dag}$\and
    Ankush Khandelwal$^\dag$\and
    Xiaowei Jia\thanks{University of Pittsburgh. \{XIAOWEI\}@pitt.edu}\and
    David Mulla$^\dag$\and
    Vipin Kumar\thanks{University of Minnesota. \{pravirat, ghosh128, khand035, mulla003, kumar001\}@umn.edu}
}
\date{}

\maketitle


\fancyfoot[R]{\scriptsize{Copyright \textcopyright\ 2024 by SIAM\\
Unauthorized reproduction of this article is prohibited}}




\begin{abstract} \small\baselineskip=9pt Accurate and timely crop mapping is essential for yield estimation, insurance claims, and conservation efforts. Over the years, many successful machine learning models for crop mapping have been developed that use just the multi-spectral imagery from satellites to predict crop type over the area of interest. However, these traditional methods do not account for the physical processes that govern crop growth. At a high level, crop growth can be envisioned as physical parameters, such as weather and soil type, acting upon the plant leading to crop growth which can be observed via satellites. In this paper, we propose Weather-based Spatio-Temporal segmentation network with ATTention (WSTATT), a deep learning model that leverages this understanding of crop growth by formulating it as an inverse model that combines weather (Daymet) and satellite imagery (Sentinel-2) to generate accurate crop maps. 
We show that our approach provides significant improvements over existing algorithms that solely rely on spectral imagery by comparing segmentation maps and F1 classification scores. Furthermore, effective use of attention in WSTATT architecture enables detection of crop types earlier in the season (up to 5 months in advance), which is very useful for improving food supply projections.
We finally discuss the impact of weather by correlating our results with crop phenology to show that WSTATT is able to capture physical properties of crop growth. 
\end{abstract}


\section{Introduction}

With increased need for food due to rapidly growing population and erratic weather patterns, accurate crop monitoring is essential for forecasting food supply and proper management of resources\cite{mora2014global}. 
A crucial aspect of crop monitoring is pixel-wise crop mapping, which assigns a crop category label to each pixel in a specified region. Accurate and timely crop maps can facilitate land use planning, yield estimation, pest management, and the evaluation of sustainable management practices and conservation efforts\cite{alami2023crop}. Various data-driven approaches, ranging from simple regression-based models \cite{julien2009yearly,olthof2007mapping} to complex deep-learning models\cite{8518619,7891032,talukdar2020land}, have been proposed to build crop maps. A vast majority of these approaches rely solely on satellite imagery sources to generate crop maps for their study region \cite{statt,teimouri2019novel,alami2023crop}. 

\begin{figure}[t]
    \centering
    \includegraphics[width = \linewidth]{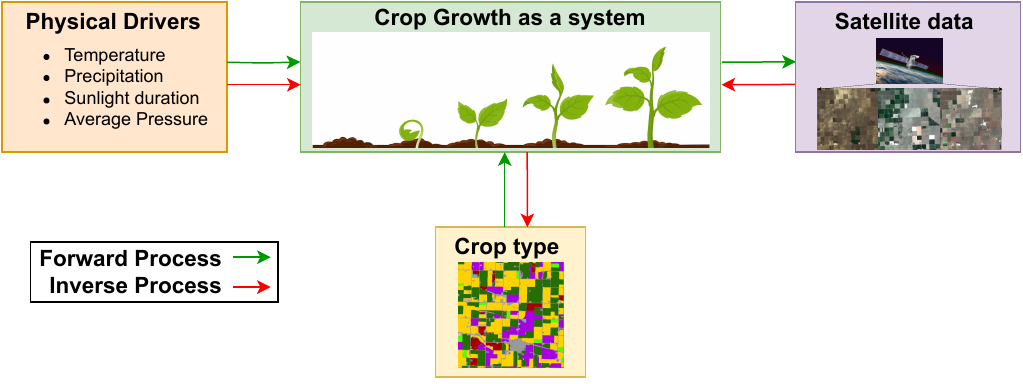}
    \vspace{-0.8cm}
    \caption{A diagrammatic representation of the physics point of view understanding of crop growth}
    \label{fig:crop_system}
    \vspace{-0.8cm}
\end{figure}

However, from a physics point of view, crop growth signal observed by satellite images is a result of complex interplay between type of crop, weather patterns, soil, management practices, etc. This view is also referred to as forward modeling of crop growth (shown as green arrows in Figure \ref{fig:crop_system}). 
Since, weather is a major source of heterogeneity (same crop can grow differently due to variation in weather and same weather pattern impact crops differently), 
we aim to leverage this forward modeling strategy by including weather data into deep learning model for the task of crop type mapping. Such an approach has the potential to offer new insights into understanding and predicting crop dynamics and improve generalizability over space and time. In physics-based models (that simulate the forward modeling of crop growth), crop-type is often considered as one of the initial conditions (or simulation parameters, knobs, etc) because the response of a system to weather drivers is conditioned on these parameters. Such a problem setting occurs in many other applications. Consider the example of predicting stream flow of a watershed, where a similar weather pattern can lead  to different streamflow depending on the watershed characteristics (such as forest cover, soil type, terrain,etc.)~\cite{ghosh2022robust}. 

Since our goal is to estimate the crop-type label, we cannot use this forward model formulation directly. Specifically, in our case, weather drivers are available as well as a proxy of crop growth (observed via satellite imagery) but the key initial condition, i.e., crop type is missing. To address this issue, we propose to use an inverse modeling~\cite{ghosh2022robust} formulation where hidden relationships between weather and crop-growth are learned to infer the initial condition (crop-type) (red arrows in Figure \ref{fig:crop_system}).  
In particular, we present WSTATT (Weather-based Spatio-Temporal segmentation networks with ATTention), a deep learning model that combines the spatio-temporal satellite and weather data with attention to give accurate pixel-wise segmentation maps for a given region. The key idea in WSTATT is to create embeddings that capture inherent relationships between weather and crops to improve classification performance.
We show the efficacy of WSTATT for crop mapping task over the state-of-the-art model, STATT \cite{statt}, that relies only on satellite imagery and was shown to be better than various spatiotemporal methods such as 3DCNN\cite{ji20183d}, ConvLSTM\cite{xingjian2015convolutional}, and CALD\cite{jia2019spatial} for crop mapping. Specifically, we show that WSTATT not only performs well on estimating crop labels, but also it can assign these labels without seeing data for the full year. This early prediction capability enables timely intervention, yield prediction, pest management, and risk mitigation.
We further show that our attention module effectively combines weather and satellite imagery to captures key discriminative periods among different crops. Finally, we discuss the impact of using weather data in prediction for each crop class by correlating the results to their phenologies.

Our contributions can be summarized as follows:
\vspace{-.02cm}
\begin{itemize}
    \item We develop an inverse modeling-based deep learning approach that uses both spatiotemporal weather and satellite data for pixel-wise crop type mapping.
    \item We show that our proposed approach outperforms the current state-of-the-art on both year-end prediction and early-prediction for crop mapping. 
    \item We provide visual analysis of attention weights to show the influence of the inverse modeling approach in effectively identifying discriminative timeframes.
    \item We qualitatively analyse the relationship between model performance and crop phenology to show the model's ability to learn meaningful patterns.
    \item We release the code \footnotemark{1}\footnotetext{\href{https://drive.google.com/drive/folders/1h43Ltes5CX_E3YZQEecB0rWcvr2Fc29f?usp=sharing}{\underline{Code}}} used in this work to promote reproducibility. 
\end{itemize}

\section{Related Work}

\subsection{Land Cover Mapping}
Various land cover mapping methods have been developed that use satellite imagery as their main input, ranging from simple regression techniques \cite{julien2009yearly,olthof2007mapping} to more complex deep learning approaches \cite{8518619,7891032,talukdar2020land}. Many of these methods involve using convolutional networks to extract spatial features \cite{hu2018deep,mahdianpari2018very}, while others utilize spatiotemporal methods to take advantage of the spatiotemporal nature of satellite data \cite{zhao2021evaluation,teimouri2019novel,statt}. These methods often involve combining recurrent networks \cite{hochreiter1997long,schuster1997bidirectional} with traditional convolution networks, and some even incorporate data from multiple satellites to address issues such as clouds and missing data\cite{blickensdorfer2022mapping}. Using these methods, many land cover datasets have been created, covering large spatial extents\cite{helber2019eurosat,yu2014multi,xia2023openearthmap,tseng2021cropharvest,m2019semantic}. One notable example is CALCROP21 \cite{calcrop21}, a large-scale dataset that provides pixel-wise information on both major and minor crops in California. 
Recently \cite{russwurm2023end} proposed a recurrent neural network based method for early prediction for crop classification using two output heads, one for each classification and stopping probability. They were able to show that their model was able to provide class labels sooner, and with less data. However, their test data was from the same year as that of training, which leads to the question of whether their probability head would carry over to a future year, as the spectral signatures, frequency of data and management practices would all change. For early prediction in years different from the training year,\cite{capliez2023temporal} suggested to use target domain data in an adversarial fashion to adapt their model and help in target domain classification. In our work, we show that inclusion of weather data inherently provides adaptation capability.


\subsection{Weather based Remote Sensing}
Crop growth is affected by weather, but its application in crop mapping is limited. Previous studies used weather data to correct their classification approaches~\cite{foerster2012crop} while training if the crop growth was inconsistent with meteorological signals. More recent research~\cite{dash2021classification} incorporated macronutrient information and weather data to predict suitable crop types using an SVM-based method. Self-organizing map-based methods were also used to forecast suitable crop growth and weather~\cite{mohan2018deep}. In a recent study~\cite{nieto2021integrated}, a combination of weather data, field information, and satellite data was used to track the growth stages of maize and provide decision-making support for maize fields. However, this study assumed knowledge of the specific crop type, which are not available in our setting. Another study such as ~\cite{addimando2022deep} briefly experimented with incorporating weather and satellite data in field crop type mapping. Although this study also used deep learning with satellite and weather data, it has several key differences from our approach. The authors propose selectively dropping weather time steps to ensure compatibility with their deep learning approach, whereas WSTATT uses recurrent networks to preserve temporal information across all timestamps. Second, this method only incorporates temporal relationships (and ignores spatial relationships) unlike WSTATT, even though it has been shown that spatio-temporal methods perform well compared to purely temporal approaches~\cite{statt}. Third, their method also does not allow for early prediction, due to the lack of variability in temporal input sequence length. Finally, they could not assess the impact of weather on their experiments, claiming that their test set was not rich enough. Weather data has been commonly used in crop-related tasks such as selecting suitable crops for maximizing crop yield~\cite{jain2020machine}, forecasting crop yield~\cite{schwalbert2020mid,milesi2022crop}, and assessing crop yield~\cite{cai2010integrating,supit2012assessing}. This usage shows that weather data has promising potential for crop-related tasks.

\section{Method}

\subsection{Problem Setting}
In this paper, we consider crop mapping as a semantic segmentation task, where we aim to assign a particular class label to each pixel of an image dataset. During training, we have the following data sources:
\begin{itemize}
    \item Satellite image time-series $S = [S^1, \dots, S^{Ts}]$, where each $S^t \in \mathbb{R}^{\text{$L_s$}\times \text{$B_s$}\times \text{$C_s$}}$ is a satellite image of size $(\text{$L_s$},\text{$B_s$})$ at time $t$ with $\text{$C_s$}$ multi-spectral channels.
    \item Weather image time-series $W = [W^1, \dots, W^{Tw}]$, where each $W^t \in \mathbb{R}^{\text{$L_w$}\times \text{$B_w$}\times \text{$C_w$}}$ is a weather data image of size $(\text{$L_w$},\text{$B_w$})$ at time $t$ with $\text{$C_w$}$ multi-spectral channels. Note that the temporal frequency of the satellite image time series and weather data can be different.
    \item Labels $Y \in {\{0,1\}}^{\text{L}\times \text{B}\times \text{V}}$ in one-hot representation, where $\text{V}$ is the number of classes. The objective is to predict one label for each pixel given the time series of satellite and weather data over a year. 
\end{itemize}

\begin{figure*}[!t]
    \centering
    \includegraphics[width = 0.9\linewidth]{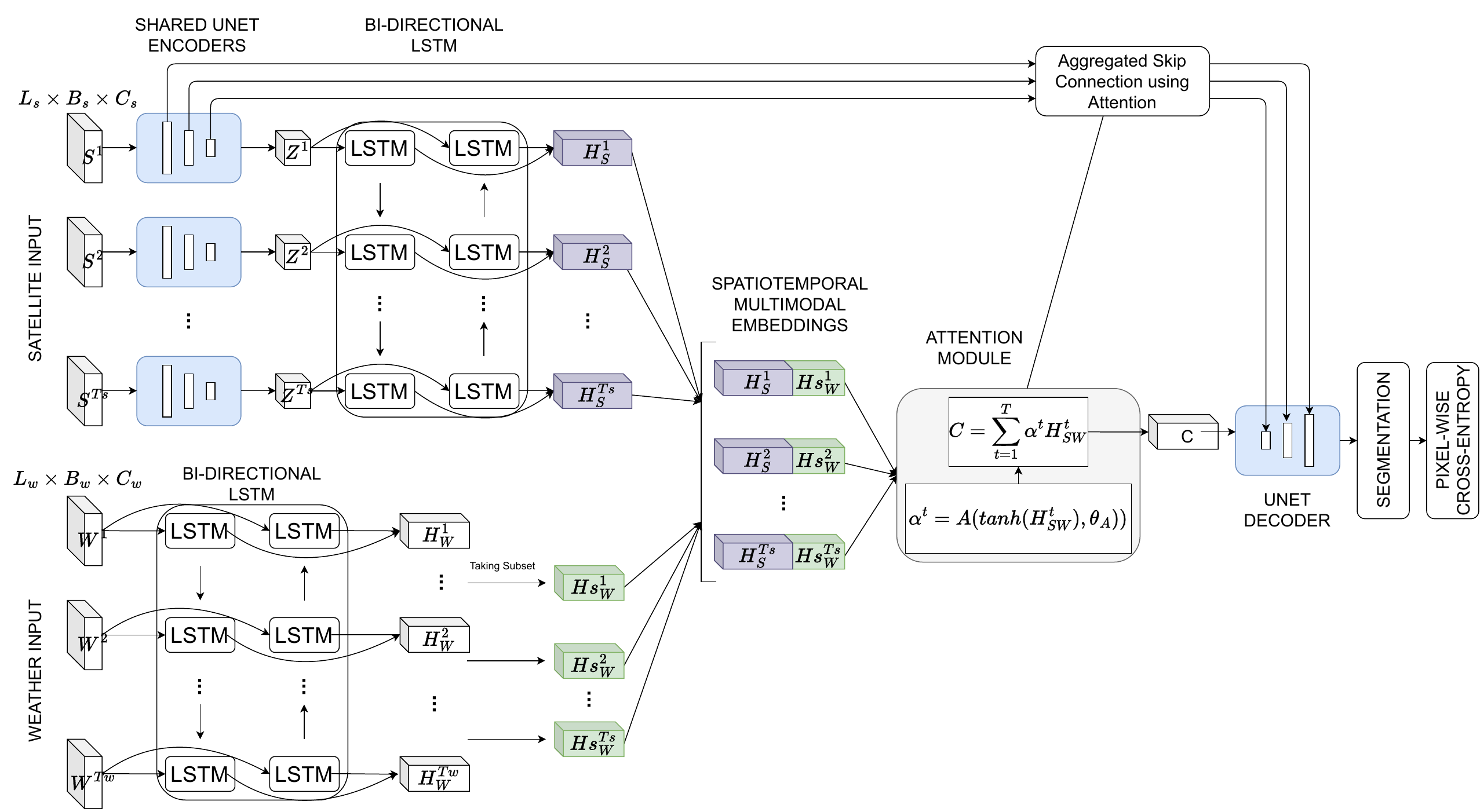}
    \vspace{-0.3cm}
    \caption{A diagrammatic representation of our proposed deep learning architecture WSTATT (Weather based Spatio-Temporal segmentation networks with ATTention) }
    \vspace{-0.6cm}
    \label{fig:wstatt}
\end{figure*}

\vspace{-0.5cm}
\subsection{Deep Learning model}
Our deep learning model is inspired by Spatio-Temporal segmentation networks with ATTention (STATT) \cite{statt}, a recent state-of-the-art segmentation model for land cover mapping. Like STATT, our model uses an encoder-decoder setup with temporal attention to predict a pixel-wise segmentation map,
However, the key difference is our formulation of crop mapping task as an inverse problem, using weather data as an additional input. 
A diagrammatic representation of our proposed deep learning architecture, WSTATT, can be seen in Figure \ref{fig:wstatt}. The architecture has two separate encoders: one for the satellite data $E_s(\,\cdot\,;\theta_{Es})$ and one for the weather data $E_q(\,\cdot\,;\theta_{Eq})$. This is followed by an attention module $A(\,\cdot\,;\theta_{A})$, which uses both encoders' outputs. Finally, the decoder $D(\,\cdot\,;\theta_{D})$ is used to create the maps.

The satellite data encoder $E_s(\,\cdot\,;\theta_{Es})$ consists of convolutional and LSTM layers to capture spatial context and temporal dynamics, respectively. The convolutional layers encode each satellite image $S^t$ to $Z^t$, and create a time series of spatial embeddings $Z = [Z^1, .., Z^{Ts}]$ 
The parameters in the convolution layers are shared across multiple images in the input time series. To capture the temporal dependencies within the spatial embeddings, this series is passed into a Bidirectional LSTM to get the hidden states $H_{ij} = [H^1_{ij}, .., H^T_{ij}]$, where T denotes the number of timestamps.
The LSTM is governed by a set of equations, given below, which use the previous hidden state $H_{ij}^{t-1}$ and cell state $C_{ij}^{t-1}$ to generate the current hidden state $H_{ij}^{t}$.

\vspace{-0.4cm}
\begin{align}
    \begin{rcases}
        F_{ij}^t &= \sigma (W_H^FH_{ij}^{t-1} + W_Z^FZ_{ij}^{t})\\
        I_{ij}^t &= \sigma (W_H^IH_{ij}^{t-1} + W_Z^IZ_{ij}^{t})\\
        O_{ij}^t &= \sigma (W_H^OH_{ij}^{t-1} + W_Z^OZ_{ij}^{t})\\
        G_{ij}^t &= \text{tanh}(W_H^GH_{ij}^{t-1} + W_Z^GZ_{ij}^{t})\\
        C_{ij}^t &= F_{ij}^t\odot C_{ij}^{t-1} + I_{ij}^t\odot G_{ij}^t\\
        H_{ij}^t &= O_{ij}^t \odot \text{tanh}(C_{ij}^t)
    \end{rcases} {i,j}\in \left(\text{H}', \text{W}'\right)
    \label{eq:Bi-LSTM}
\end{align}

Each crop class has a different growing pattern across the entire year. However, any two crop classes are generally most distinguishable only during a certain time window, which we refer to as the discriminative period. To effectively capture this period, examining the forward and backward directions across all available timestamps is necessary, leading to the use of Bidirectional LSTM in our method. The Bidirectional LSTM produces two hidden states for each timestamp, which we concatenate to create the final series of hidden representations $H_S = [H_S^1, \dots, H_S^T]$.

Similarly, the weather data encoder .$E_q(\,\cdot\,;\theta_{Eq})$, consists of a Bidirectional LSTM to encode the weather data into its respective series of hidden states $H_W = [H_W^1, \dots, H_W^T]$. Compared with satellite images, weather data are often collected at a higher temporal frequency (e.g., daily) but at a coarser spatial resolution. As a result, in many cases, a training patch will have only a single weather pixel within the patch bounds. Hence, we do not use convolutional layers for encoding the weather data, and use only a Bidirectional LSTM encoder resulting in a hidden embedding for each timestamp of the weather data. As a result, the length of the weather embeddings list ($H_W$) is equal to the number of timestamps of weather data present. Since weather data and satellite data are available at different temporal frequency, the vectors $H_W$ and $H_S$ will be of unequal lengths. To address this issue, we select weather data embeddings at the same time intervals as the satellite data's frequency. Specifically, if the satellite data is available every 15 days and weather data is available daily, we would take every 15$^{th}$ weather data embedding to match the frequency of satellite data. Since each hidden state is created using previous hidden states, we still use information from all weather data even though we take a subset of embeddings. So even if, for example, the 15$^{th}$ day has noisy weather readings, since the 15$^{th}$ weather embedding is created using this noisy reading and all previous weather readings, it will still contain relevant information. We also resorted to equally spaced selection subsetting as opposed to averaging-based subsetting because we did not want dynamic changes in weather to be lost in the averaging. After this equally spaced selection of embeddings, we are left with a subset of weather data embeddings $Hs_W = [Hs_W^1, \dots, Hs_W^{Ts}]$, the same length as $H_S$. Now, to match the spatial dimensions of $H_S$, each entry in $Hs_W$ is resampled spatially using the same value (as there is only one embedding value for each $\text{H}\times \text{W}$ of satellite data. Futhermore, we concatenate $H_S$ and $Hs_W$ along the channel dimension to create $H_{SW} = [H_{SW}^1, \dots, H_{SW}^{Ts}]$. This series represents the spatiotemporal multimodal embeddings of both the satellite images and the weather data, thus enabling the inverse modeling approach by forcing all subsequent components in the method to use both modalities jointly. 

This multimodal embedding series is then passed onto the attention network $A(\,\cdot\,;\theta_{A})$, which assigns a weight for each timestamp dynamically. The attention weight for each timestamp ranges in (0,1) and sums up to 1 over all the timestamps. The weight represents the importance of data at each timestamp towards the final goal of crop mapping. We used a single-layer feed-forward network as the attention layer for our implementation. This layer helps the model focus on which timestamps are more discriminative for any given class and can also help reducing the effect of issues such as cloud cover blockage or missing data. The series $H_{SW}$ is aggregated temporally using these attention weights $(\alpha^t)$ to form the final embedding $C_{SW}$.

This attention-aggregated embedding is then sent into the decoder. The decoder $D(\,\cdot\,;\theta_{D})$ is a set of convolutional layers, similar to the UNET deconvolution approach. Since the input to the decoder is a multimodal attention aggregated series, the decoder is forced to learn the relationship between these embeddings and the final crop map, which is the idea behind the inverse modeling paradigm. Similar to STATT, we also use aggregated skip connections using attention, using the attention weights $(\alpha^t)$ at every step of the decoding. Finally, we use a linear layer followed by softmax to get the pixel-wise class probabilities. The model can be trained using pixel-wise cross-entropy loss.

\section{Dataset and Experimental Setup}
\subsection{Study Region}
\begin{figure}[!t]
    \centering
    \includegraphics[width = 0.95\linewidth]{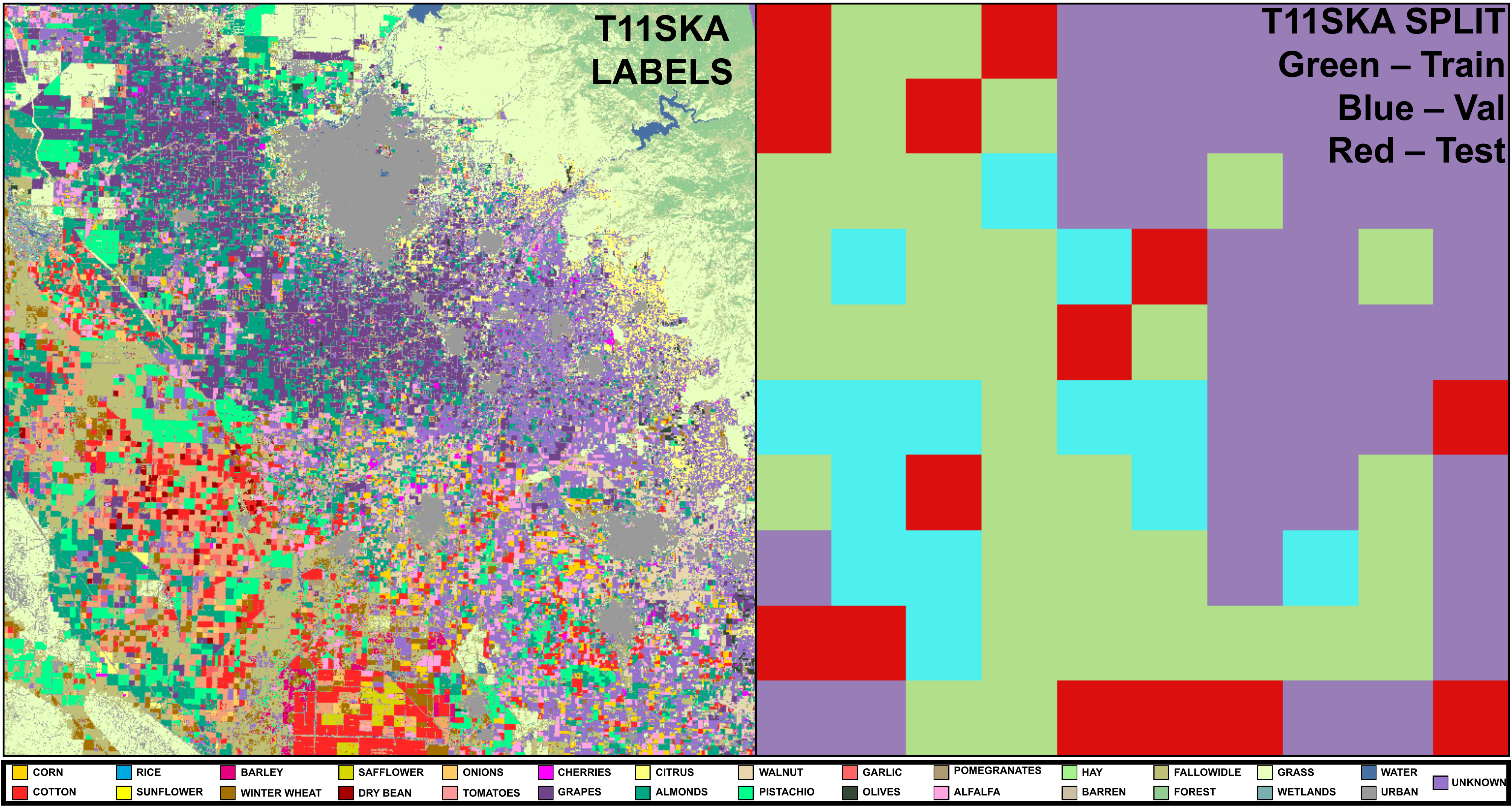}
    \vspace{-0.3cm}
    \caption{Visualisation of T11SKA Sentinel Tile CDL labels and grid split used in our experiments. Regions in Purple represent Unknown and were not used in analysis}
    \label{fig:region_analysis}
    \vspace{-0.5cm}
\end{figure}

We selected the crop belt in Central Valley of California as our study region (100km x 100km) due to the richness in crops grown and its weather variability. Specifically, we chose the regions bounded by the T11SKA Sentinel-2 Tile, an area with over 30 crop classes. A visualization of the crop classes in this region can be seen in Figure~\ref{fig:region_analysis}. These class labels are taken from the USDA Crop Data Layer (also done by other works such as CALCROP21~\cite{calcrop21}).

\subsection{Satellite Imagery}
We use multi-spectral satellite imagery from Sentinel-2 satellites \cite{sentinel2} for our experiments. Specifically, we downloaded data for year 2018 from the COPERNICUS/S2\_SR 
\footnote{\url{https://developers.google.com/earth-engine/datasets/catalog/COPERNICUS_S2_SR##bands}} collection on Google Earth Engine.
The Sentinel-2 data product has 13 spectral bands at different spatial resolutions of 10, 20, and 60 meters. We leave out the atmospheric bands (Band 1, 9, and 10) of 60 meters resolution (due to coarse resolution) and re-sample all the remaining bands to 10 meters using the nearest neighbor method. 
Due to irregular temporal sampling of Sentinel-2 imagery across years we create a multi-spectral mosaic at a 15-day interval while considering cloud filters at every timestamp. This resulted in a final dataset of shape 24 x 10 x 10980 x 10980 for a year, where dimensions represents number of composite images, channels, rows, and columns respectively with each row and columns pixel denoting a 10m spatial length (Thus each satellite image's  height and width is 109800m). Each band is normalised with its respective maximum and minimum.
One can obtain similar mosaic sets for any year, such as 2019 or 2020. 

\subsection{Weather Data}
We used Daymet dataset as our source of weather data. Specifically, we downloaded daily weather data from NASA/ORNL/DAYMET\_V4 
\footnote{\url{https://developers.google.com/earth-engine/datasets/catalog/NASA_ORNL_DAYMET_V4}} collection on Google Earth Engine.
The Daymet product has seven bands namely, duration of the daylight period, daily total precipitation, incident shortwave radiation flux density, snow water equivalent,  maximum 2-meter air temperature, minimum 2-meter air temperature, and average partial pressure of water vapor. Daymet is available daily at 1km spatial resolution. However, we found that downloading the data at this resolution led to a lot of missing data, so we download all bands of data daily at a resolution of 10 km for our study region. We then resample weather data (using nearest neighbor interpolation method) to match the spatial extents with our multi-spectral satellite data and normalise each band with its respective minimums and maximums. This resulted a final shape of 365 x 7 x 10 x 10 for a year, where dimensions represents number of timestamps, channels, rows, and columns respectively  with each row and columns pixel denoting a 10km spatial length.

\subsection{Labels}
We get our labels from the Cropland Data Layer (CDL) \footnote{\url{https://nassgeodata.gmu.edu/CropScape/}} provided by the United States Department of Agriculture (USDA). USDA annually releases the CDL \cite{cdl} which is a publicly available land-cover classification map for the entire country at a 30-meter resolution that includes major and minor crop commodities. Since CDL is available at 30 m resolution, we resample it to 10m using nearest neighbor to match our satellite imagery dataset. There are over 200 class types in CDL, with some irrelevant to our crop mapping task, so we follow a class combination scheme similar to that performed in CALCROP21\cite{calcrop21} by combining the pixel labels to one of 33 classes namely \{Corn, Cotton, Rice, Sunflower, Barley, Winter Wheat, Safflower, Dry Beans, Onions, Tomatoes, Cherries, Grapes, Citrus, Almonds, Walnut, Pistachio, Garlic, Olives, Pomegranates, Alfalfa, Hay, Barren, Fallow and Idle, Deciduous Forests, Evergreen forest, Mixed Forests, Clover and wildflower, Shrubland, Grass, Woody wetlands, Herbaceous Wetlands, Water, Urban\}. In addition, We also have an {Unknown} class to denote those pixels that do not belong to our 33 classes of interest. Therefore, for our study region, the final labels are of shape 34 x 10980 x 10980 for each year.

\vspace{-0.2cm}
\subsection{Implementation details}
Like CALCROP21 \cite{calcrop21}, we adopt a grid-based splitting of the study region to ensure spatial auto-correlation does not negatively impact our evaluation. Specifically, we divide our study region into 100 grids of size 10kmx10km each (equivalent to 1098 x 1098 pixels). Furthermore, we eliminate grids that do not contain more than 50\% pixels belonging to crop classes. 
It is also known that CDL has issues with noise and incorrect labels at the boundaries of crop fields. To address this, we perform one level of morphological erosion, i.e remove one layer of pixels at object boundaries to reduce the effect of label uncertainty at farm boundaries. We further remove small connected components (contiguous set of pixels belonging to the same class) to reduce the effect of noisy labels in the training process, please refer to \cite{calcrop21} for diagrammatic representation of this process. Since we follow a grid-wise scheme for the satellite data, we also split the weather data into this grid scheme. This results in one weather reading per day for each grid, which is a reasonable approximation, as it can be expected for a 10km x 10km region to have very similar weather due to large spatial auto-correlation in weather patterns.

\section{Results}
The main objective of this paper is to show the benefit of bringing in weather data for crop mapping compared to methods that use only satellite data. As mentioned, for the satellite-only based approach, we chose STATT \cite{statt} to be compared against our weather-based model. 


\textbf{Training Set}: Following the removal of grids due to lack of crop cover, we were left with 62 grids for the T11SKA region in 2018. From these 62 grids, We randomly selected 34 grids for training, 14 for validation, and remaining 14 for testing. We keep the training and validation sets across the experiments the same.  As mentioned before, each grid is of size 1098 x 1098 pixels, thus signifying a large spatial region and also each grid is consistent across time. The distribution of these grids can be seen in Figure \ref{fig:region_analysis}. By using the grid based splitting of our study region, we avoid spatial auto-correlation between pixels (nearby pixels tend to be of same type) to negatively affect our evaluation. Furthermore, we exclude 30 pixels on each side of each grid to further remove the effect of spatial auto-correlation in our analysis. Both WSTATT and STATT were trained using these 34 grids, and the best model was chosen using the 14 validation grids. Both models were trained using learning rate of 0.0001, 50 epochs and using cross-entropy loss. 

\subsection{Year End Prediction}
In this section, we discuss the predictive performance of experiments where entire year's data is available for making the crop type prediction, i.e predicting the crop map at the end of the year with all 12 months of data available. 
First, we evaluate STATT and WSTATT on 14 test grids (denoted by the red boxes in Fig. \ref{fig:region_analysis}) using data from the same year as the training data (2018). Specifically, we compare their F1 scores for different crop classes while ignoring classes with less than 100,000 pixels from this evaluation. We do not compare accuracy as it does not give a complete picture of correctness of each method, due to imbalance in classes for which F1 score is better tuned.
WSTATT performed better than STATT with an overall improvement of 0.02 in average F1 score, with major improvements in Corn (0.08) and Walnuts (0.04).  
We further evaluate these algorithms in a more difficult setting where data for 14 test grids come from different years (2019 and 2020). The idea is to assess models' ability to generalize across time. For this setting, we observe a 0.15 increase in average F1 score for year 2019 using WSTATT compared to STATT, and an increase of 0.12 for year 2020 (Please refer to Table \ref{Tab:avgf1} for average and classwise comparison of F1 Score)
Major improvements were seen in Almonds (0.31 increase in 2019 and 0.18 increase in 2020) and also in Alfalfa (0.25 increase in both years). 
As the results suggest, WSTATT is able to learn robust discriminative features among crop classes by leveraging the relationship between weather and crop-growth. 


\begin{table*}[t!]
\centering
\caption{Classwise comparison of WSTATT and STATT in terms F1 Score for Year end predictions across various years. The numbers in bold correspond to the best across each experiment.}
\footnotesize
\begin{tabular}{|c|cc|cccc|cccc|}
\hline
              & \multicolumn{6}{c|}{Year end Prediction}                                                                                                   \\ \hline
Test Set      & \multicolumn{2}{c|}{T11SKA 2018}                       & \multicolumn{2}{c|}{T11SKA 2019}              & \multicolumn{2}{c|}{T11SKA 2020}  \\ \hline
Class         & STATT           & \multicolumn{1}{c|}{WSTATT}          & STATT  & \multicolumn{1}{c|}{WSTATT}          & STATT           & WSTATT          \\ \hline
Corn          & 0.718           & \multicolumn{1}{c|}{\textbf{0.8014}} & 0.5761 & \multicolumn{1}{c|}{\textbf{0.7002}} & 0.7376          & \textbf{0.7425} \\
Cotton        & 0.958           & \multicolumn{1}{c|}{\textbf{0.9583}} & 0.8888 & \multicolumn{1}{c|}{\textbf{0.8951}} & 0.8866          & \textbf{0.8983} \\
Winter\_Wheat & \textbf{0.7276} & \multicolumn{1}{c|}{0.7135}          & 0.2166 & \multicolumn{1}{c|}{\textbf{0.3516}} & \textbf{0.7083} & 0.6028          \\
Tomatoes      & 0.8727          & \multicolumn{1}{c|}{\textbf{0.8921}} & 0.7790 & \multicolumn{1}{c|}{\textbf{0.7986}} & 0.7114          & \textbf{0.7893} \\
Grapes        & \textbf{0.8691} & \multicolumn{1}{c|}{0.8682}          & 0.6719 & \multicolumn{1}{c|}{\textbf{0.8038}} & 0.7424          & \textbf{0.7857} \\
Citrus        & 0.7842          & \multicolumn{1}{c|}{\textbf{0.8007}} & 0.7814 & \multicolumn{1}{c|}{\textbf{0.8471}} & 0.6054          & \textbf{0.7881} \\
Almonds       & 0.8197          & \multicolumn{1}{c|}{\textbf{0.8435}} & 0.4615 & \multicolumn{1}{c|}{\textbf{0.7728}} & 0.5320          & \textbf{0.7117} \\
Walnut        & 0.8164          & \multicolumn{1}{c|}{\textbf{0.8537}} & 0.5187 & \multicolumn{1}{c|}{\textbf{0.7252}} & 0.4190          & \textbf{0.6250} \\
Pistachio     & 0.8447          & \multicolumn{1}{c|}{\textbf{0.8778}} & 0.4649 & \multicolumn{1}{c|}{\textbf{0.7261}} & 0.3189          & \textbf{0.6418} \\
Alfalfa       & 0.7605          & \multicolumn{1}{c|}{\textbf{0.7892}} & 0.4985 & \multicolumn{1}{c|}{\textbf{0.7447}} & 0.5397          & \textbf{0.7827} \\ \hline
Average       & 0.8171          & \multicolumn{1}{c|}{\textbf{0.8398}} & 0.5857 & \multicolumn{1}{c|}{\textbf{0.7365}} & 0.6201          & \textbf{0.7368} \\ \hline
\end{tabular}
\vspace{-0.5cm}
\label{Tab:avgf1}
\end{table*}

\begin{table*}[!h]
\centering
\caption{Comparison of WSTATT and STATT in terms F1 Score over various early prediction settings on test grids in 2019 and 2020. Note that both models are trained using only data from 2018. The numbers in bold correspond to the best across each experiment.}
\footnotesize
\begin{tabular}{|ccccccccc|}
\hline
\multicolumn{9}{|c|}{T11SKA 2019 Early Prediction}                                                                                                                                                                               \\ \hline
\multicolumn{1}{|c|}{Data Provided} & \multicolumn{2}{c|}{6 MONTHS}                          & \multicolumn{2}{c|}{8 MONTHS}                 & \multicolumn{2}{c|}{10 MONTHS}                & \multicolumn{2}{c|}{12 MONTHS}    \\ \hline
\multicolumn{1}{|c|}{Crop Class}    & STATT           & \multicolumn{1}{c|}{WSTATT}          & STATT  & \multicolumn{1}{c|}{WSTATT}          & STATT  & \multicolumn{1}{c|}{WSTATT}          & STATT           & WSTATT          \\ \hline
\multicolumn{1}{|c|}{Corn}          & \textbf{0.2747} & \multicolumn{1}{c|}{0.0819}          & 0.3606 & \multicolumn{1}{c|}{\textbf{0.6883}} & 0.5043 & \multicolumn{1}{c|}{\textbf{0.7091}} & 0.5761          & \textbf{0.7002} \\
\multicolumn{1}{|c|}{Cotton}        & 0.0064          & \multicolumn{1}{c|}{\textbf{0.6156}} & 0.1934 & \multicolumn{1}{c|}{\textbf{0.8483}} & 0.7590 & \multicolumn{1}{c|}{\textbf{0.8943}} & 0.8888          & \textbf{0.8951} \\
\multicolumn{1}{|c|}{Winter\_Wheat} & 0.0500          & \multicolumn{1}{c|}{\textbf{0.3850}} & 0.2000 & \multicolumn{1}{c|}{\textbf{0.2785}} & 0.2112 & \multicolumn{1}{c|}{\textbf{0.3042}} & 0.2166          & \textbf{0.3516} \\
\multicolumn{1}{|c|}{Tomatoes}      & 0.1618          & \multicolumn{1}{c|}{\textbf{0.3469}} & 0.5890 & \multicolumn{1}{c|}{\textbf{0.7763}} & 0.7485 & \multicolumn{1}{c|}{\textbf{0.8204}} & 0.7790          & \textbf{0.7986} \\
\multicolumn{1}{|c|}{Grapes}        & 0.0134          & \multicolumn{1}{c|}{\textbf{0.2306}} & 0.0727 & \multicolumn{1}{c|}{\textbf{0.5806}} & 0.4015 & \multicolumn{1}{c|}{\textbf{0.6654}} & 0.6719          & \textbf{0.8038} \\
\multicolumn{1}{|c|}{Citrus}        & 0.5473          & \multicolumn{1}{c|}{\textbf{0.8148}} & 0.6593 & \multicolumn{1}{c|}{\textbf{0.8041}} & 0.7348 & \multicolumn{1}{c|}{\textbf{0.8285}} & 0.7814          & \textbf{0.8471} \\
\multicolumn{1}{|c|}{Almonds}       & 0.0757          & \multicolumn{1}{c|}{\textbf{0.2293}} & 0.1794 & \multicolumn{1}{c|}{\textbf{0.7100}} & 0.3078 & \multicolumn{1}{c|}{\textbf{0.7503}} & 0.4615          & \textbf{0.7728} \\
\multicolumn{1}{|c|}{Walnut}        & \textbf{0.2788} & \multicolumn{1}{c|}{0.2308}          & 0.3341 & \multicolumn{1}{c|}{\textbf{0.6913}} & 0.4889 & \multicolumn{1}{c|}{\textbf{0.7619}} & 0.5187          & \textbf{0.7252} \\
\multicolumn{1}{|c|}{Pistachio}     & 0.0000          & \multicolumn{1}{c|}{\textbf{0.1142}} & 0.0006 & \multicolumn{1}{c|}{\textbf{0.4340}} & 0.0198 & \multicolumn{1}{c|}{\textbf{0.4682}} & 0.4649          & \textbf{0.7261} \\
\multicolumn{1}{|c|}{Alfalfa}       & 0.2676          & \multicolumn{1}{c|}{\textbf{0.5762}} & 0.3393 & \multicolumn{1}{c|}{\textbf{0.6364}} & 0.4300 & \multicolumn{1}{c|}{\textbf{0.7310}} & 0.4985          & \textbf{0.7447} \\ \hline
\multicolumn{1}{|c|}{Average}       & 0.1676          & \multicolumn{1}{c|}{\textbf{0.3625}}& 0.2928 & \multicolumn{1}{c|}{\textbf{0.6448}}& 0.4606          & \multicolumn{1}{c|}{\textbf{0.6933}}& 0.5857 & \multicolumn{1}{c|}{\textbf{0.7365}} \\ \hline
\multicolumn{9}{|c|}{T11SKA 2020 Early Prediction}                                                                                                                                                                               \\ \hline
\multicolumn{1}{|c|}{Data Provided} & \multicolumn{2}{c|}{6 MONTHS}                          & \multicolumn{2}{c|}{8 MONTHS}                 & \multicolumn{2}{c|}{10 MONTHS}                & \multicolumn{2}{c|}{12 MONTHS}    \\ \hline
\multicolumn{1}{|c|}{Crop Class}    & STATT           & \multicolumn{1}{c|}{WSTATT}          & STATT  & \multicolumn{1}{c|}{WSTATT}          & STATT  & \multicolumn{1}{c|}{WSTATT}          & STATT           & WSTATT          \\ \hline
\multicolumn{1}{|c|}{Corn}          & \textbf{0.6267} & \multicolumn{1}{c|}{0.3181}          & 0.6395 & \multicolumn{1}{c|}{\textbf{0.6694}} & 0.6970 & \multicolumn{1}{c|}{\textbf{0.7596}} & 0.7376          & \textbf{0.7425} \\
\multicolumn{1}{|c|}{Cotton}        & 0.1491          & \multicolumn{1}{c|}{\textbf{0.6658}} & 0.3435 & \multicolumn{1}{c|}{\textbf{0.8740}} & 0.4530 & \multicolumn{1}{c|}{\textbf{0.9076}} & 0.8866          & \textbf{0.8983} \\
\multicolumn{1}{|c|}{Winter\_Wheat} & \textbf{0.6477} & \multicolumn{1}{c|}{0.6182}          & 0.6438 & \multicolumn{1}{c|}{\textbf{0.6707}} & 0.6454 & \multicolumn{1}{c|}{\textbf{0.6531}} & \textbf{0.7083} & 0.6028          \\
\multicolumn{1}{|c|}{Tomatoes}      & \textbf{0.4693} & \multicolumn{1}{c|}{0.2796}          & 0.5801 & \multicolumn{1}{c|}{\textbf{0.7259}} & 0.6364 & \multicolumn{1}{c|}{\textbf{0.7906}} & 0.7114          & \textbf{0.7893} \\
\multicolumn{1}{|c|}{Grapes}        & 0.0413          & \multicolumn{1}{c|}{\textbf{0.1541}} & 0.2857 & \multicolumn{1}{c|}{\textbf{0.4810}} & 0.6023 & \multicolumn{1}{c|}{\textbf{0.6762}} & 0.7424          & \textbf{0.7857} \\
\multicolumn{1}{|c|}{Citrus}        & 0.5321          & \multicolumn{1}{c|}{\textbf{0.7276}} & 0.6058 & \multicolumn{1}{c|}{\textbf{0.7563}} & 0.5317 & \multicolumn{1}{c|}{\textbf{0.7948}} & 0.6054          & \textbf{0.7881} \\
\multicolumn{1}{|c|}{Almonds}       & 0.3764          & \multicolumn{1}{c|}{\textbf{0.6488}} & 0.3664 & \multicolumn{1}{c|}{\textbf{0.6705}} & 0.3325 & \multicolumn{1}{c|}{\textbf{0.6774}} & 0.5320          & \textbf{0.7117} \\
\multicolumn{1}{|c|}{Walnut}        & 0.4810          & \multicolumn{1}{c|}{\textbf{0.6106}} & 0.4577 & \multicolumn{1}{c|}{\textbf{0.8146}} & 0.2063 & \multicolumn{1}{c|}{\textbf{0.7945}} & 0.4190          & \textbf{0.6250} \\
\multicolumn{1}{|c|}{Pistachio}     & 0.0002          & \multicolumn{1}{c|}{\textbf{0.2685}} & 0.0033 & \multicolumn{1}{c|}{\textbf{0.3595}} & 0.0153 & \multicolumn{1}{c|}{\textbf{0.3960}} & 0.3189          & \textbf{0.6418} \\
\multicolumn{1}{|c|}{Alfalfa}       & 0.2859          & \multicolumn{1}{c|}{\textbf{0.6322}} & 0.3793 & \multicolumn{1}{c|}{\textbf{0.6903}} & 0.4958 & \multicolumn{1}{c|}{\textbf{0.7610}} & 0.5397          & \textbf{0.7827} \\ \hline
\multicolumn{1}{|c|}{Average}       & 0.3610         & \multicolumn{1}{c|}{\textbf{0.4924}}& 0.4305 & \multicolumn{1}{c|}{\textbf{0.6712}}& 0.4616         & \multicolumn{1}{c|}{\textbf{0.7211}}& 0.6201 & \multicolumn{1}{c|}{\textbf{0.7368}} \\ \hline
\end{tabular}
\vspace{-0.5cm}
\label{Tab:earlypred2}
\end{table*}

\subsection{Early Prediction}

In this section we evaluate the performance of the models for early prediction, i.e obtaining the crop map without entire year's data. 
Both STATT and WSTATT have attention modules allowing dynamic timestamp weight allocation followed by aggregation. Hence, even with partial year's data, the attention module can assign weights to available timesteps and provide prediction probabilities allowing for early prediction at various time scales (6 months, 8 months, etc). 

For this experiment, we vary the amount of data available for prediction as 6, 8, 10, and 12 months, where the year starts from January. Table \ref{Tab:earlypred2} shows the early prediction results for 2019 and 2020. As we can see, WSTATT performs significantly better than STATT in the early prediction, for both years 2019 and 2020 which were not considered during training.  

With just eight months of data in the test regions over both years, WSTATT consistently outperforms STATT with major differences in Cotton, Almonds, Pistachio and Grapes. This trend extends to 10 month predictions as well.
Another important factor to note is that, though performance improves for both models with more data, WSTATT reaches its best score values much faster. For example, from the Cotton scores in 2019, we observe that with just eight months of data, WSTATT can achieve good accuracy, and with four more months of data, the score increases by 0.05. However, in the same timeframe, STATT's eight-month score is poor and shows a good value only when the full year's data is provided. This indicates that WSTATT is using the weather data to make earlier predictions effectively using the partial year's data. 

\begin{figure}[t]
    \centering
    \includegraphics[width = \linewidth]{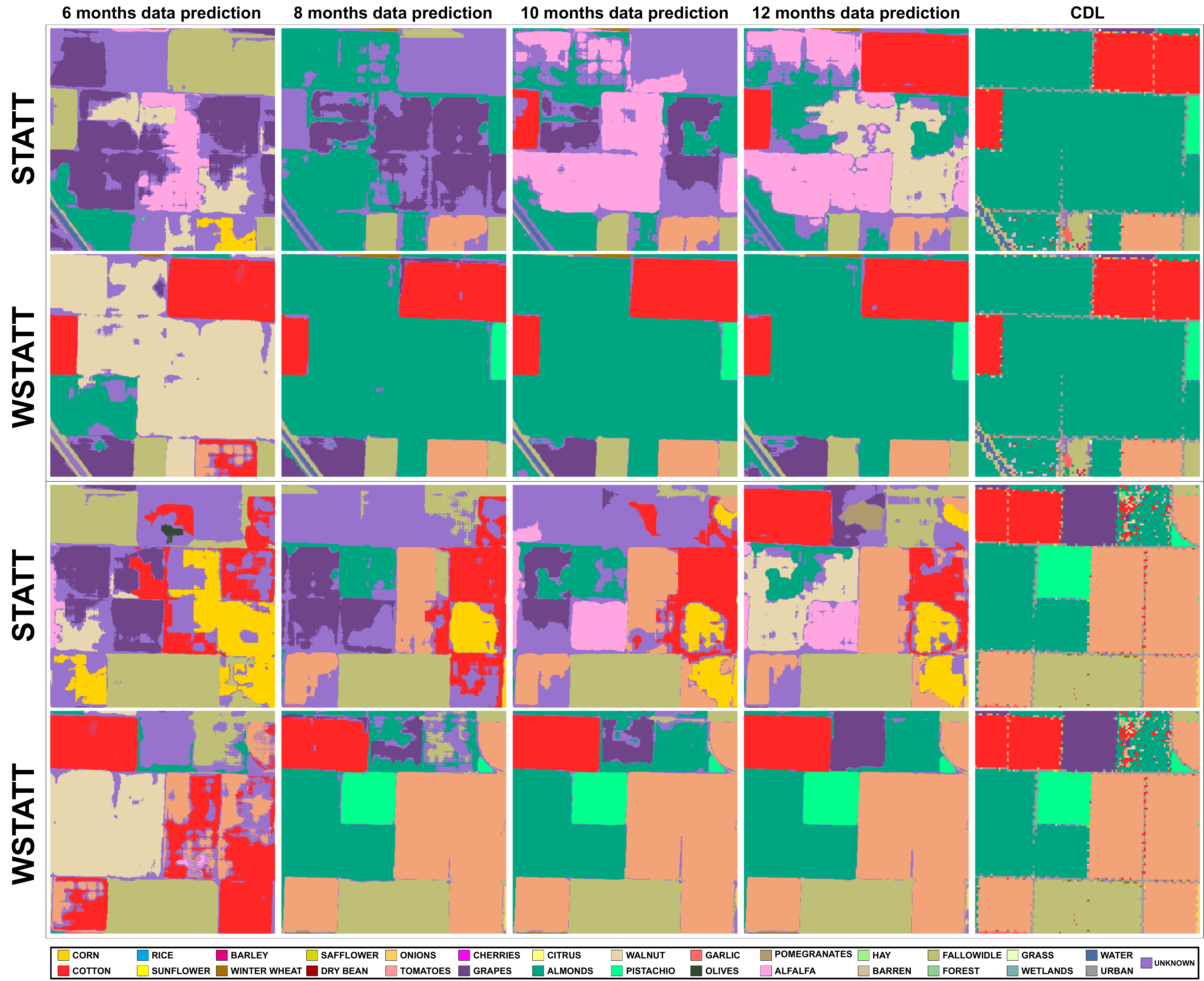}
    \vspace{-0.8cm}
    \caption{Cases where weather based model (WSTATT) gets a good prediction well before the end of the year}
    \label{fig:early_pred_8months}
\end{figure}

To provide visual analysis, we show prediction maps from 2019 for certain regions. Figure~\ref{fig:early_pred_8months} illustrates two such cases where the prediction using only eight months of data is close to the ground truth of CDL (shown in the last column). Although some corrections are made with more data, the majority of the patch is accurate. However, with the same amount of data, STATT cannot provide a good prediction and even with more data provided is unable to reach an accurate prediction. 

\begin{figure}[t]
    \centering
    \includegraphics[width = \linewidth]{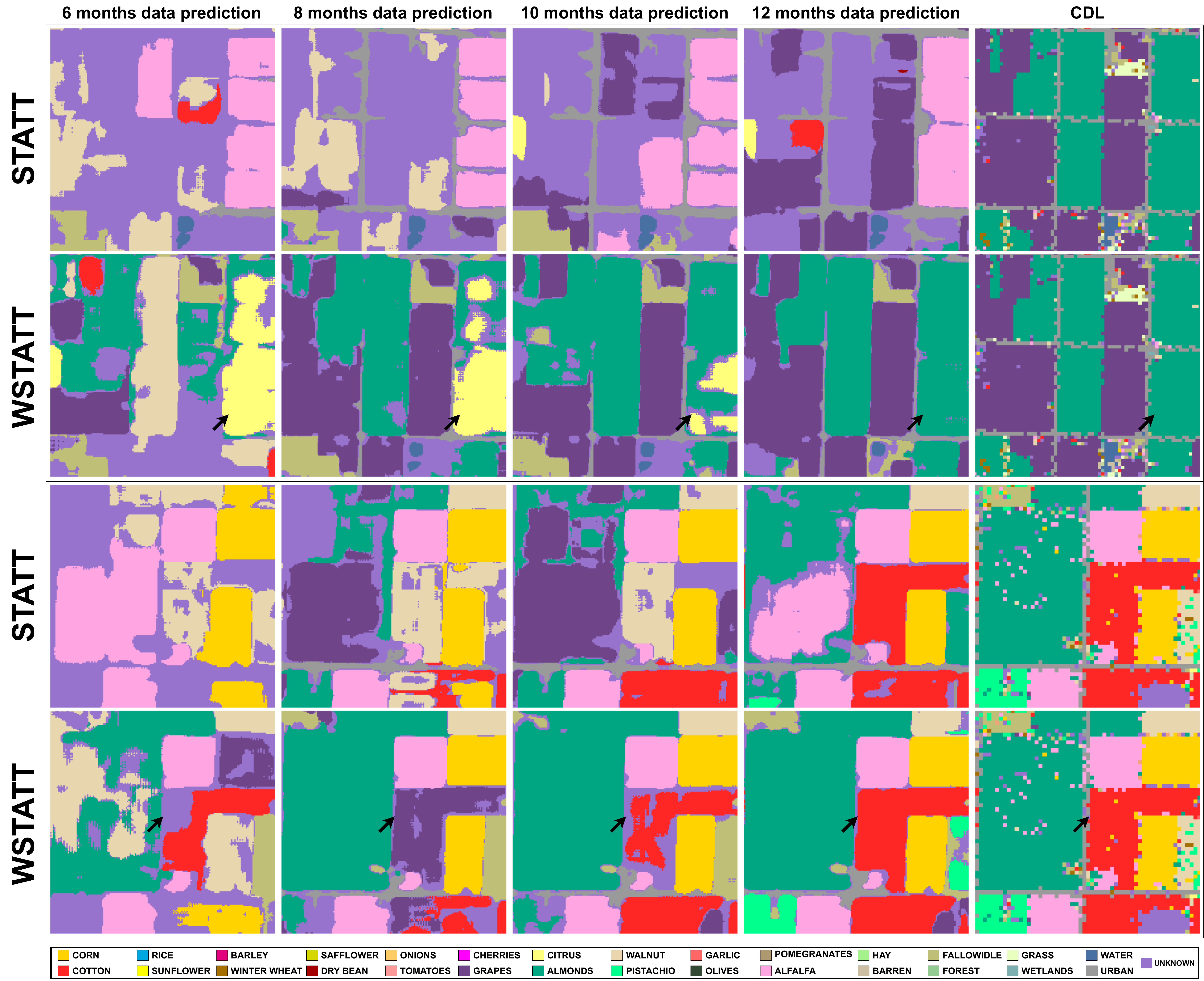}
    \vspace{-0.8cm}
    \caption{Cases where as more data is provided the prediction of incorrect fields progressively gets better (Please look at fields denoted by arrow marks)}
    \label{fig:early_pred_prog}
\end{figure}

Figure~\ref{fig:early_pred_prog} shows two more cases from 2019 where WSTATT can correct some fields as more data is provided, but STATT is unable to do so. In the top pair of Figure~\ref{fig:early_pred_prog}, WSTATT has some errors in an Almond's field (dark green) at the six-month prediction but improves and eventually becomes perfect by the end of the year. However, STATT cannot correct itself and predicts Alfalfa for that field, even with the full year's data. Similarly, in the bottom pair, WSTATT corrects two fields, Almonds(dark green) and Cotton(red), as more data is given, with Almonds being corrected faster when compared to Cotton. In that same time frame, STATT is only able to reach a correct prediction for Cotton with the full year's data, but not for Almond's field despite having full data.

Through our experiment, we have found that the weather-based model can adjust to different temporal scales and produce accurate maps well before the end of the year, reaching even higher accuracy by the end of the year. This is a significant advancement in crop mapping since the CDL map for a given year is usually provided a month after the year ends, which would be around late January or early February of the next year. Our method can provide the map as early as August of the same year, which can benefit downstream tasks that rely on robust crop maps such as yield prediction.

\vspace{-.3cm}

\subsection{Attention Analysis}
The previous sections show that the weather-based model outperforms the satellite-only approach which can be attributed to the integration of the inverse modeling scheme in WSTATT. 
While both approaches utilize a feed forward attention module that assigns a weight to each timestep, STATT relies only on satellite data but WSTATT uses weather data as well. In this section, we examine the impact of including weather data on the attention module.

Some key factors about a crop are its planting date, growth pattern, biomass at maturity, harvest date, crop residue, and leaf shedding. They all can play a role in discriminating crop classes but not all of them are consistent across years. For example, planting and harvesting dates can change over years. Algorithms that identify features that are more generalizable such as the growth pattern, timing of crop maturity, etc. are more likely to perform better across different years. Since weather is tightly coupled with crop growth and maturity, it enables WSTATT to focus more on generalizable features instead of over-fitting on noisy features such as crop residue after harvesting, planting date etc.

\begin{figure}[t]
    \centering
    \includegraphics[width = 0.95\linewidth]{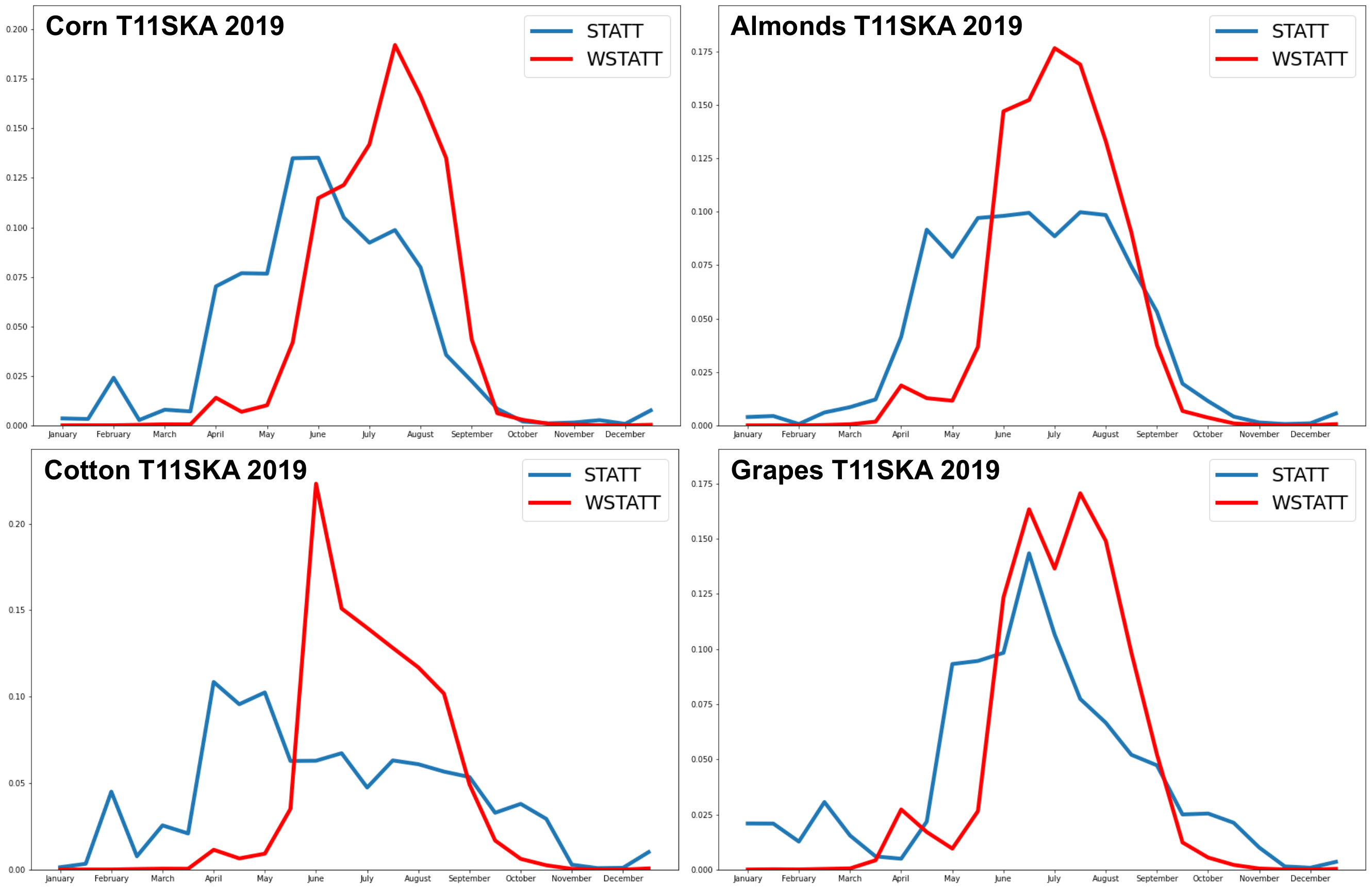}
    \vspace{-0.5cm}
    \caption{A diagrammatic representation of the attention weights for each crop class in 2019}
    \vspace{-0.3cm}
    \label{fig:attention_analysis}
\end{figure}

Figure \ref{fig:attention_analysis} shows attention weights assigned by both algorithms to each timestep for four crop classes. Each subplot displays a different crop, indicated in the top left corner, with the x-axis representing the timeframe and the y- axis showing the weightage for the corresponding timestamp. To ensure attention weights are representative of that class, we selected a patch that belong completely to that class. As we can see, attention weights are very different from both algorithms.  WSTATT weights are much concentrated on few timesteps compared to STATT which suggest better understanding of discriminative periods for different classes.
For instance, WSTATT focused on June to September for almonds, which was the easiest period to distinguish them from other crops. In contrast, STATT attention scores were almost equal from April to September. Similarly, for Corn, STATT focused on the period from April to July, which could cause confusion with winter wheat, while WSTATT paid attention to July to September. 


\vspace{-0.2cm}
\section{Discussion on Impact of Weather}

In this section, we provide domain insights into the trends observed in F1 scores for the two algorithms as reported in Table \ref{Tab:earlypred2}. 
Since Citrus and Alfalfa maintain their leaves throughout the year, both algorithms are providing good predictive performance even from 6 months onwards with WSTATT significantly better for these two crops even at 6 months. Grapes, a perennial crop, are harvested in September or October and shed their leaves during winter. WSTATT is quicker than STATT in detecting this, with a notable increase in performance of 8 month prediction as opposed to 10 month prediction for STATT. Almonds, pistachios, and walnuts are also perennial crops, but they shed their leaves later in the year, around October to December, causing a jump in their performance around the eight months timestamp. Pistachio, being the last to shed its leaves, shows a boost in performance only on the last timestamp. 

Tomatoes have different planting times but are harvested between July and October. Therefore, their predictive performance shows a significant increase at eight months for WSTATT. Similarly, Cotton is typically planted in April, but it reaches good vegetative cover later in the year, resulting in a sharp increase in predictive performance for WSTATT but not for STATT. Corn is usually planted in May, but its leaf area index is still low after six months, making it difficult to classify accurately. At this time, winter wheat is almost at the end of its growth cycle, and some fields may have crop residue, which can be confused with corn. However, since winter wheat's growth cycle is complete, Table \ref{Tab:earlypred2} shows that its predictive performance does not improve significantly beyond the six-month prediction, regardless of the data provided for both methods over the years.


The results indicate that including weather data is useful for identifying harvest periods more quickly and accurately capturing intervals of leaf loss. This can be attributed to the fact that inclusion of weather data allows for the model to quickly discern the difference between crop classes. A key point to note is that while higher frequency satellite imagery might offer improvements in early prediction, but the key factor determining early prediction capability is the timing of the discriminative period for a crop, irrespective of the temporal resolution. 
\vspace{-.1cm}

\section{Conclusion and Future Work}    
Our work introduces a new method for crop mapping that considers crop growth as a system influenced by physical drivers such as weather and soil type. We developed an inverse modeling approach using satellite imagery and weather data to provide accurate pixel-wise crop maps for a given area of interest. Our weather-based method outperforms traditional satellite-only methods for year end and early prediction tasks. Our early prediction allows for robust crop maps to be available to farmers and the public much sooner than the current standard (upto 5 months in advance). We found that including weather embeddings improved attention scores and allowed the model to focus on the most discriminative period. 

For future work, we plan to include a generative forward modeling process in combination with the current inverse framework. Currently, our focus was on evaluating the performance over different time periods. We plan to extend the study region, and assess how weather data can enable generalization across space by comparing against existing domain adaptation techniques. 
Another future work direction is to include more physical parameters in the deep learning approach, such as soil information and management practices. 

In conclusion, this work showed the immense benefit of bringing weather data into pixel-wise crop mapping and we believe that this work will convince remote sensing practitioners to use such an approach in their tasks, replacing the traditional methods of relying solely on satellite imagery for land cover mapping tasks.


\section{Acknowledgments}
This work was funded by the NSF awards 2313174 and 1838159. Access to computing facilities was provided by the Minnesota Supercomputing Institute.


\bibliographystyle{plain}
\bibliography{bibliography}
\vspace{-1cm}

\end{document}